\relax
\documentclass[letterpaper]{article} 
\usepackage{aaai19}  
\usepackage{times}  
\usepackage{helvet}  
\usepackage{courier}  
\usepackage{url}  
\usepackage{graphicx}  

\usepackage{amsmath} 
\usepackage{amssymb}
\usepackage{multirow}
\usepackage{subfigure}
\usepackage{makecell}
\usepackage{caption}
\usepackage{lipsum}
\newcommand{\tabincell}[2]{\begin{tabular}{@{}#1@{}}#2\end{tabular}}

\title{Segregated Temporal Assembly Recurrent Networks \\for Weakly Supervised Multiple Action Detection}
\author{
Yunlu Xu\textsuperscript{1}\thanks{Authors contribute equally. Zhang did this work during an internship in Hikvision Research Institute. }
Chengwei Zhang\textsuperscript{2}\footnotemark[1]
Zhanzhan Cheng\textsuperscript{13}\thanks{Corresponding author.} Jianwen Xie\textsuperscript{1} Yi Niu\textsuperscript{1} Shiliang Pu\textsuperscript{1} Fei Wu\textsuperscript{3}\\
\textsuperscript{1}Hikvision Research Institute, China;  ~~~\textsuperscript{2}Shanghai Jiaotong University, China; ~~~\textsuperscript{3}Zhejiang University, China\\
\{xuyunlu,chengzhanzhan,jianwen.xie,niuyi,pushiliang\}@hikvision.com;~
cwzhang@sjtu.edu.cn;~wufei@cs.zju.edu.cn
}

\begin{document}
\maketitle

\begin{abstract}
This paper proposes a segregated temporal assembly recurrent (STAR) network for weakly-supervised multiple action detection. The model learns from untrimmed videos with only supervision of video-level labels and makes prediction of intervals of multiple actions.
Specifically, we first assemble video clips according to class labels by an attention mechanism that learns class-variable attention weights and thus helps the noise relieving from background or other actions.
Secondly, we build temporal relationship between actions by feeding the assembled features into an enhanced recurrent neural network.
Finally, we transform the output of recurrent neural network into the corresponding action distribution.
In order to generate more precise temporal proposals, we design a score term called segregated temporal gradient-weighted class activation mapping (ST-GradCAM) fused with attention weights.
Experiments on THUMOS'14 and ActivityNet1.3 datasets show that our approach outperforms the state-of-the-art weakly-supervised method, and performs at par with the fully-supervised counterparts.
\end{abstract}

\section{Introduction}
\noindent
Multiple action detection, which aims at localizing temporal intervals of actions and simultaneously identifying their categories in videos, is a fundamental problem in video understanding. Many existing works \cite{Shou2016Temporal,Zhao2017Temporal,Shou2017CDC,Xu2017R,yang2018exploring,Chao2018Rethinking,Lin2018BSN,Alwassel2018ECCV} make efforts to address this problem in a supervised manner, where the algorithms rely on fully labeled data (i.e., videos with precise annotations of the starting and ending frames of actions). However, such supervised methods are prohibitively impractical in real applications, since frame-level annotations are substantially time-consuming and expensive. Therefore, learning to detect temporal action from untrimmed videos remains a crucial and challenging problem in video understanding.

A few explorations based solely on video-level annotations have exemplified the weakly supervised temporal action detection.
UntrimmedNet \cite{wang2017untrimmednets} proposes to learn a selection module for detecting important segments, and later STPN \cite{Nguyen2017Weakly} conquers the single-label limitation by introducing temporal class activation maps (T-CAM) trained with cross-entropy loss.
To address the issue that performing localization via thresholds may not be robust to noises in class activation maps, AutoLoc \cite{Shou_2018_ECCV} directly predicts temporal boundary and proposes a Outer-Inner-Contrastive loss to provide the desired segment-level supervision. W-TALC \cite{Paul_2018_ECCV} introduces the Co-Activity Similarity Loss and jointly optimizes it with the cross-entropy loss for the weakly-supervised temporal action detection. However, {the interference and relationship among actions in a video hitherto have not been concerned}.

In reality, a video in general describes multiple actions occurring in a complex background.
Intuitively, a desired video descriptor should have two characteristics: (1) refraining from the interference of other unrelated actions or background, and (2) enhancing the correlation among actions.

\begin{figure}[t]
\centering
	\includegraphics[width=0.48\textwidth]{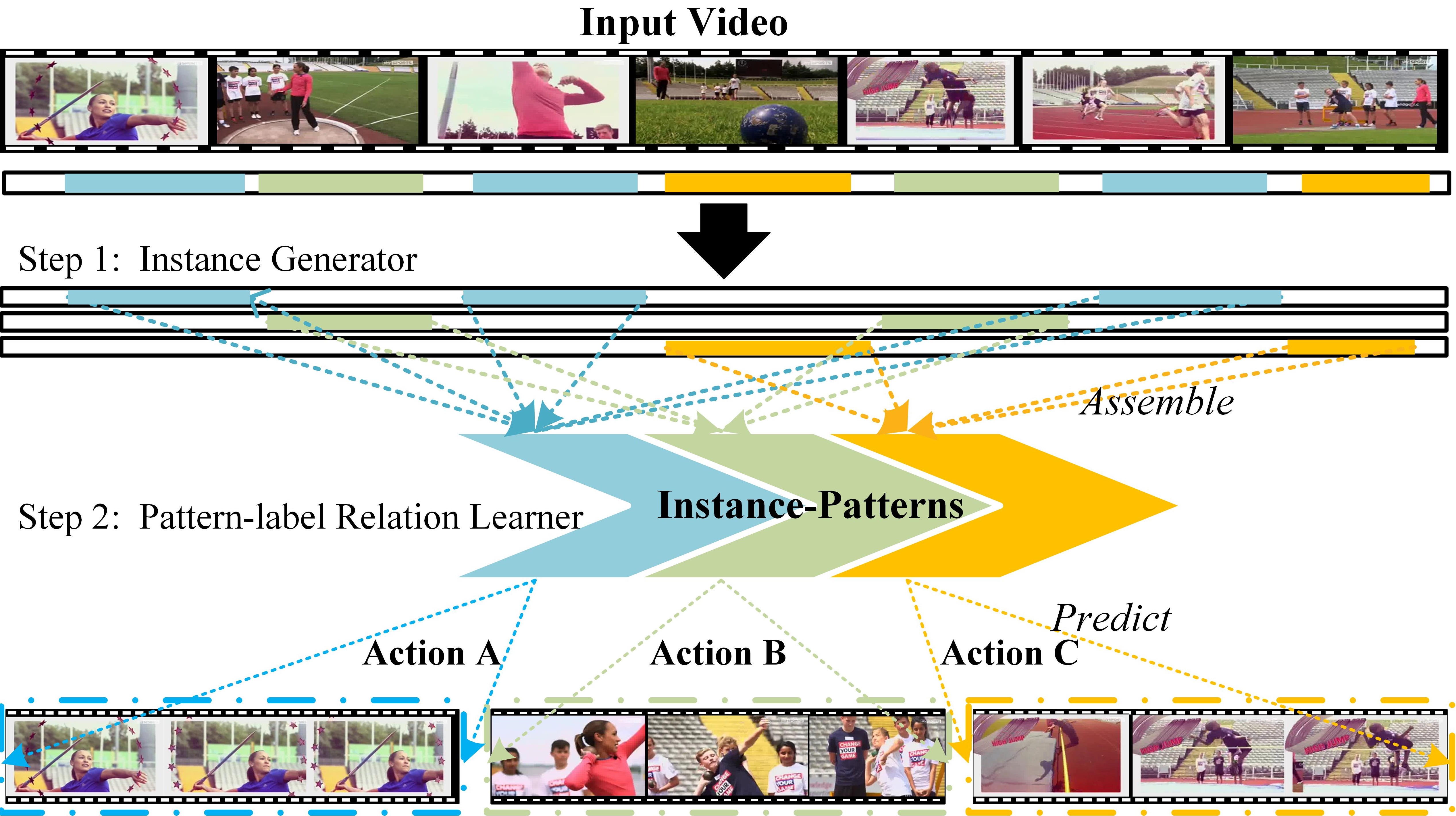}
	\caption{An illustration of the multiple action detection task from the perspective of MIML.
}
	\label{fig:intro}
\end{figure}
In this paper, we focus on the task of weakly supervised multiple action detection with only video-level labels. As illustrated in Figure \ref{fig:intro}, the task of multiple action detection in a weakly-annotated video can be regarded as a multi-instance multi-label (MIML) \cite{Zhou2008Multi} problem where an example (i.e., a video) is described by multiple instances (i.e., actions) and associated with multiple class labels (i.e., action categories).
Correspondingly, we address the weakly supervised action detection task in the following two steps. Step 1: Action assembling for multi-instance pattern generation. In order to eliminate interference from unrelated actions or complex background, we generate instance-patterns for each type of action via an action selector. As shown in Figure 1, three intervals of \emph{Action-A} (denoted in blue) are selected and integrated as an assembled feature representation \emph{Instance-pattern-A}. By this way, each instance-pattern is mapped from the corresponding type of action in the input video. Step 2: Relation learning for label generation. With assembled patterns, the corresponding instances can be directly predicted, but some correlation (\emph{e.g.,} \textit{CricketShot} always co-occurs with \textit{CricketBowling} ) generally exists in the case of multiple actions in a video. Therefore, we learn the implicit relationship between different instances by adopting a recurrent neural network \cite{hochreiter1997long}. Furthermore, the corresponding instance proposals can be activated from the learned class-variable weights.

More specifically, we propose a weakly-supervised framework for multiple temporal action detection called \textbf{S}egregated \textbf{T}emporal \textbf{A}ssembly \textbf{R}ecurrent (\emph{abbr}. STAR) network.
Firstly, we construct a well-designed attention module to learn the action \emph{assembly weights} for integrating the encoded segmented features into corresponding instance-patterns.
Then we learn the relationship between instances by adopting an enhanced recurrent neural network (RNN) for action label generating.
We also involves a \emph{repetition align mechanism} in RNN for adaptively adjusting the attention weights to generate finer action proposals. 
Finally, we design an operation term called  segregated temporal Gradient-CAM (ST-GradCAM),
which is an extension of Gradient-CAM \cite{gradcam}, to indicate feature significance for a specific action category.
We fuse the response of ST-GradCAM with the learned assembly weights for the purpose of action localization.

 The contributions of our paper are
four-fold:
(1) We reformulate the multiple action detection from a MIML perspective, i.e., extracting instance-patterns and generating action labels, which eliminates interference among unrelated action features and captures temporal dependency between multiple concurrent actions.
(2) An end-to-end framework called STAR, which includes a well-designed attention module and an enhanced RNN, is developed to be trained in a weakly supervised manner from videos with only video-level labels.
(3) We design an ST-GradCAM operation fused with class-variable assembly weights for action temporal localization.
(4) Experiments demonstrate that our weakly supervised framework achieves impressive performance on the challenging  THUMOS'14 \cite{THUMOS14} and ActivityNet1.3 \cite{heilbron2015activitynet:} datasets for action detection, comparable with those of supervised learning methods.

\section{Related Work}
\textit{Action Recognition.} The task of action recognition seeks to identify a single or multiple action labels for each video and is often treated as a classification problem. Before the era of deep learning, hand-crafted features, such as the improved dense trajectories \cite{wang2013action}, obtained outstanding performance on many benchmark datasets. Recently, there have been vast works on action recognition using convolutional neural networks (CNN). For example, a 2D CNN for large-scale video classification was first investigated in \cite{karpathy2014large-scale}, but has not achieved comparable performance with hand-crafted features. Two-stream \cite{simonyan2014two-stream} and C3D \cite{tran2015learning,tran2017convnet,carreira2017quo} networks are recent mainstreams to learn discriminative features for action
recognition.
The inception 3D (I3D) \cite{carreira2017quo} is a two-stream network based on a 3D version of Inception network \cite{ioffe2015batch},
which is commonly used as a feature encoder for action localization \cite{Nguyen2017Weakly} and dense-labeling videos \cite{piergiovanni2018learning}.

\textit{Fully Supervised Action Detection.}
Different from action recognition, action detection aims to identify the temporal intervals containing target actions.
Most existing works focus on fully-supervised approaches for that.
To capture robust video feature representation, S-CNN \cite{Shou2016Temporal} uses a multi-stage CNN for temporal action localization.
SSN \cite{Zhao2017Temporal} introduces structured temporal pyramids with decoupled classifiers for classifying actions and determining completeness.
For precise boundaries, the Convolutional-De-Convolutional (CDC) network \cite{Shou2017CDC}
 and the Temporal Preservation Convolutional (TPC) network \cite{yang2018exploring} are proposed for frame-level action predictions. {Boundary Sensitive network (BSN) \cite{Lin2018BSN} is recently proposed to locate temporal boundaries which are further integrated into action proposals.}
Furthermore, some region-based methods, e.g.,  R-C3D \cite{Xu2017R} and TAL-Net \cite{Chao2018Rethinking}, propose to generalize the methods for 2D spatial detection to 1D temporal localization.

\textit{Weakly Supervised Action Detection.} Action detection in a weakly supervised fashion has been studied by only a few works.
UntrimmedNet \cite{wang2017untrimmednets} is an end-to-end model for learning single-label action classification and action detections. Hide-and-seek \cite{Singh2017Hide} tries to force the model to see different parts of the
image and focus on multiple relevant parts of
the object beyond just the most discriminative one by randomly masking different regions of training images in each training epoch. Such a method works well for spatial object detection but is unsatisfied for the temporal action detection. STPN\cite{Nguyen2017Weakly} adopts an attention module to identify a sparse subset of key segments associated with target actions in a video, and fuse the key segments via adaptive temporal pooling.
The latest work \cite{Shou_2018_ECCV} and \cite{Paul_2018_ECCV} boost the STPN by introducing novel objective functions to separately tune coarse action boundaries and unearth co-activity relationship between videos.

\textit{Multi-instance Multi-label framework.}
Single-instance single-label \cite{krizhevsky2012imagenet} framework has led to remarkable performance.
While in the real-world settings, an example is usually composed of multiple instances, such as sentences in a text, image frames in a video and objects in an image. Multi-instance multi-label (MIML) \cite{Zhou2008Multi} framework, where an example is described by multiple instances and associated with multiple class labels, has been applied to different tasks, such as multi-label image classification \cite{Wang2016CNN}, image retrieval \cite{Zhang2018Instance}, object detection and semantic segmentation \cite{wei2017stc:,Ge2018Multi}, and sound separation \cite{Gao2018Learning}.
Our paper generalizes this framework to the action detection task where a video contains multiple actions and is associated with multiple video-level classes.

\begin{figure*}[t]
\centering
	\includegraphics[width=0.95\textwidth]{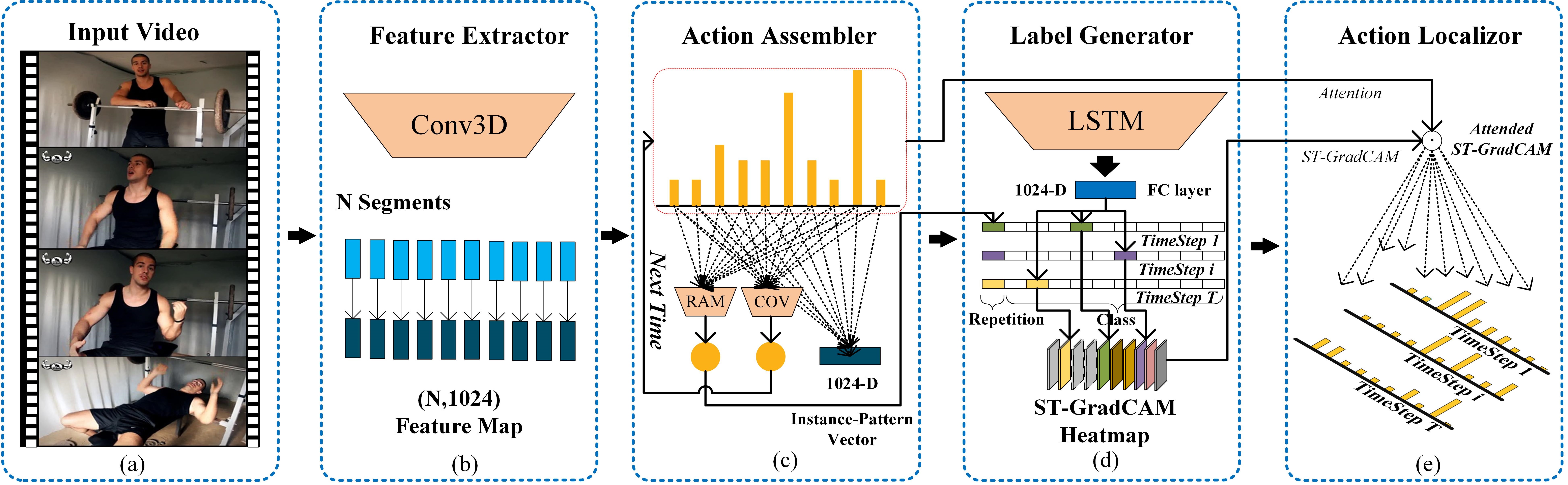}
	\caption{
        The workflow of the STAR framework. (a) The input video; (b) A pre-trained video encoder for segmental feature extraction; (c) An action assembler for generating instance-patterns, in which stage attention weights are trained with well-designed sub-modules (e.g., RAM); (d) A LSTM-based network for action label generation ; (e) A localizer for locating actions from input video, only used for inference without any training.}
	\label{fig:overview}
\end{figure*}
\section{Segregated Temporal Assembly Network}
The framework of STAR, which is shown in Figure~\ref{fig:overview}, consists of three  components:
(1) a pre-trained feature extractor for encoding a video into a sequence of segmental feature vectors, (2) an end-to-end trainable architecture that we call segregated temporal assembly network, including an action assembler and a label generator,
and (3) a well-designed action localizor for action location.
In the following, we first present the architecture of the proposed STAR in \emph{Section 3.1};
Then a well-designed action proposal mechanism is described in \emph{Section 3.2};
Finally, the training strategy of the whole network is given in \emph{Section 3.3}.


\subsection{Action Assembly and Label Generation}

\subsubsection{Stem Architecture}
\begin{figure}[h]
  \centering
  \subfigure[softmax]{
    \label{fig:subfig:a} 
    \includegraphics[width=1.4in]{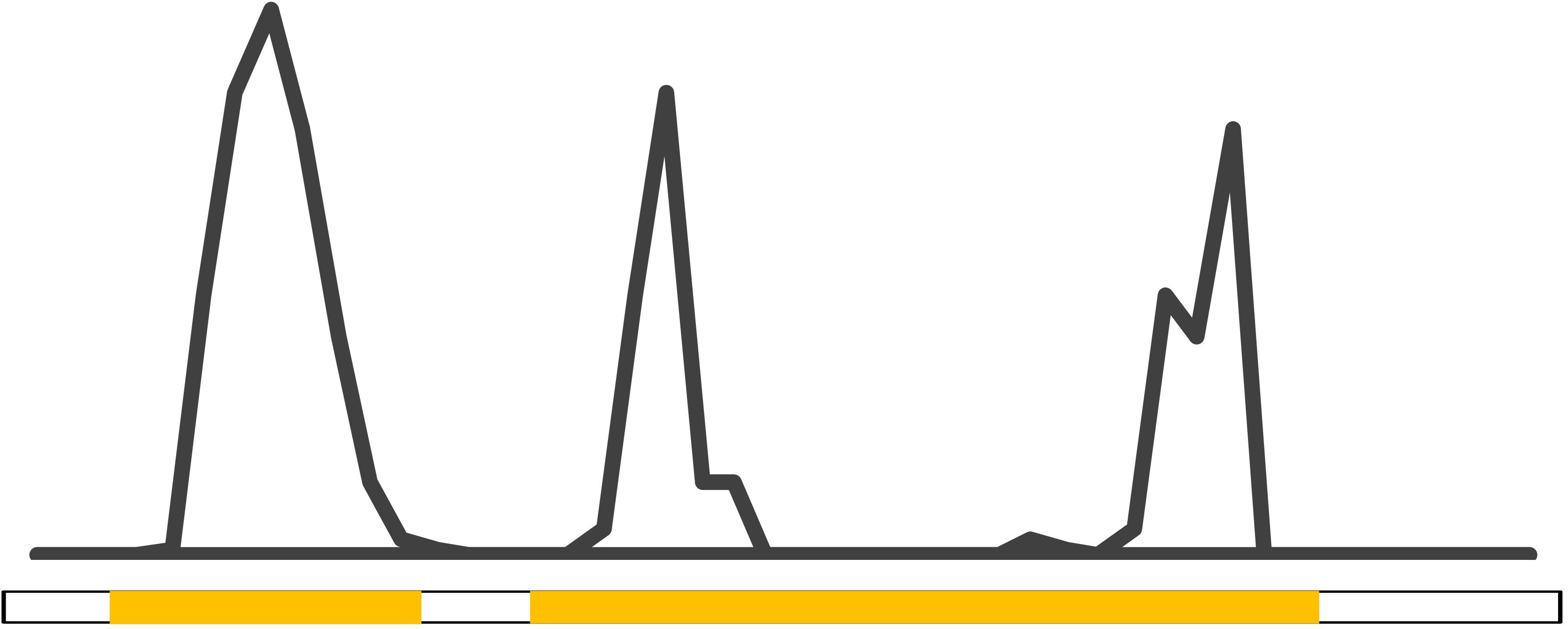}}
  \hspace{0.1in}
  \subfigure[sigmoid]{
    \label{fig:subfig:b} 
    \includegraphics[width=1.5in]{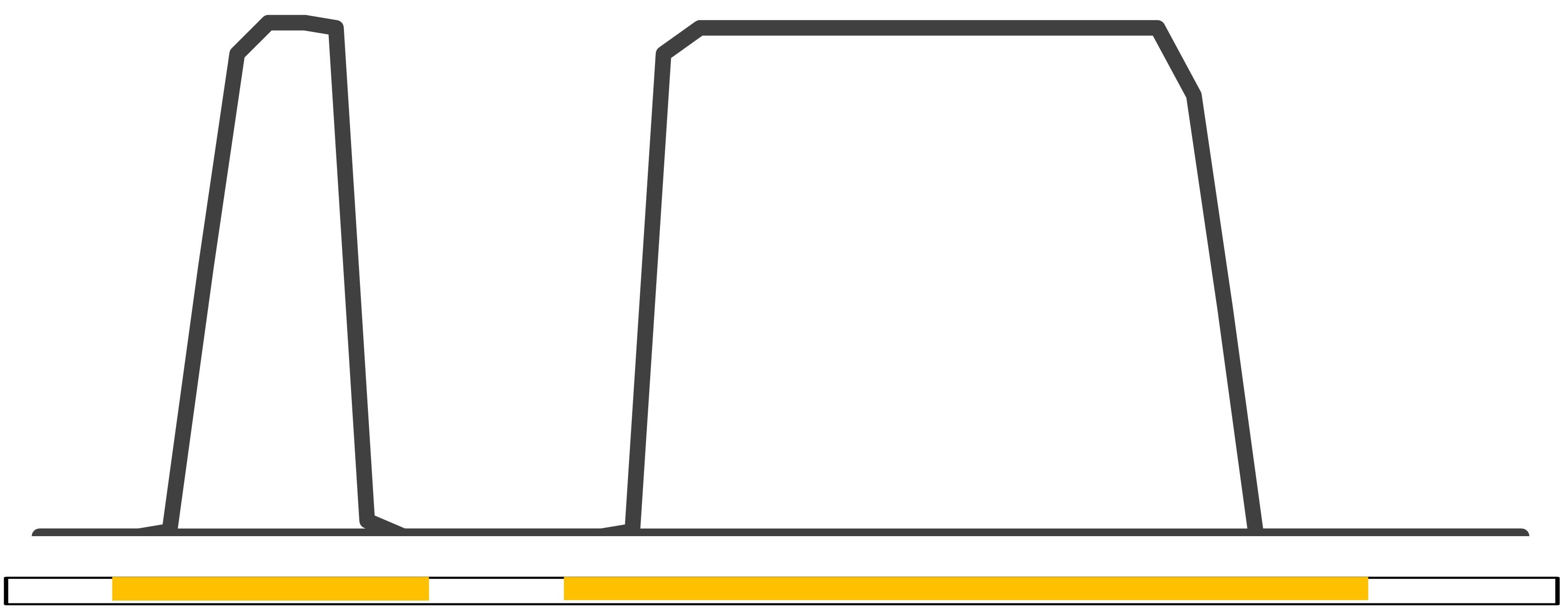}}
  \caption{Effects of different normalization functions on energy distribution for attending actions. Horizontal and vertical axes separately represent time and energy score. The ground truth of the action regions is represented by a bar, with yellow color indicating action occurrence.}
  \label{fig:subfig} 
\end{figure}

Given $N$ segments of $K$-dimensional feature vectors $S{=}\{s_1, s_2, \ldots, s_N\}  \in \mathbb{R}^{K\times N}$, which are extracted from a video $\mathcal{V}$ by a pre-trained feature extractor, we first assemble actions from $S$ into instance-patterns $X{=}\{x_1, x_2,\ldots, x_T\}  \in \mathbb{R}^{K\times T}$, where $T$ is the number of assembly actions in $\mathcal{V}$.
Then we use an RNN to build relation between the assembled actions in $X$, and further generate the action labels $y_i$ from a label set $Y=\{y_1, y_2,\ldots, y_T\}$ one-by-one.
Concretely, we first assemble actions into a specific instance-pattern at time $t$ by
\begin{equation} \label{att1}
x_{t}=\sum_{i=1}^{N}\alpha_{t,i}s_{i},
\end{equation}
where $\alpha$ is the learnt attending assembly weights over $S$.
Generally, $\alpha$ is calculated by simultaneously referring the last hidden states of RNN and glimpsing the whole input $S$ \cite{chorowski2015attention-based}.
Correspondingly, we first evaluate the energy state $e$ over $S$ by
\begin{equation} \label{att3}
e_{t,i}=v_{\alpha}\varphi(W_{\alpha}h_{t-1} + U_{\alpha}s_{i}),
\end{equation}
where $h_{t-1}$ is the hidden state of RNN at time $t\text{-}1$;
Then the energy state is further normalized by
\begin{equation} \label{att4}
\alpha_{t,i}=\sigma(e_{t,i}),
\end{equation}
where $\varphi$ and $\sigma$ are the activation function \emph{tanh} and \emph{sigmoid} respectively, and $W_\alpha$, $U_\alpha$ and $v_{\alpha}$ are learnable parameters.
Note that, the conventional attention mechanism \cite{chorowski2015attention-based} uses  \textit{softmax} function to normalize the energy distribution, which results in failures to capture those long or high-frequency actions. 
Instead of \emph{softmax}, we adopt the bounded logistic sigmoid activation function $\sigma$ on the energy distribution to deal with this issue. Figure \ref{fig:subfig} illustrates a comparison of effects of different activation functions.

After assembling actions from $S$ to $X$, we use the RNN to build relation between instance-patterns in $X$ by
\begin{equation} \label{att2}
h_{t}=RNN(h_{t-1},  y_{t-1}, x_{t}),
\end{equation}
where RNN is specified as the popular relation learning model long short-term memory (LSTM) \cite{hochreiter1997long}.
Then we further output the action probability distribution by
\begin{equation} \label{lstm7}
y_{t}=softmax({W}_o{h}_{t}),
\end{equation}
where ${W}_o$ is the learnt parameters.

Though the above process, denoted as the naive {stem} network, can generate action labels to a certain extent, it is still unsatisfactory to handle complicated video action tasks due to the following three key factors:
\begin{enumerate}
  \item \emph{Attending repetition}:
    Repetition of attending regions is a common problem for sequence-to-sequence models \cite{tu2016modeling} and is especially pronounced when generating multiple instance-patterns (see \emph{Raw Attention} in Figure \ref{fig:cov}), which goes against the purity of single pattern.
  \item \emph{Co-occurrence}:
    Unlike usual sequential applications (e.g., text reading, speech recognition, translation etc.), co-occurrence is universal in videos (\emph{e.g.,} \textit{CricketBowling} and \textit{CricketShot} always appear successively and have overlapping), which increases the challenge of video tasks. 
  \item \emph{Trivial action missing}:
    Some repetitive but inapparent action features are usually shielded by the corresponding prominent representative patterns due to the lack of frame-level annotations, leading to failures of trivial action detection.
\end{enumerate}

\subsubsection{Permissive Coverage Constraint}
\begin{figure}[h]
  \centering
  \includegraphics[width=3.2in]{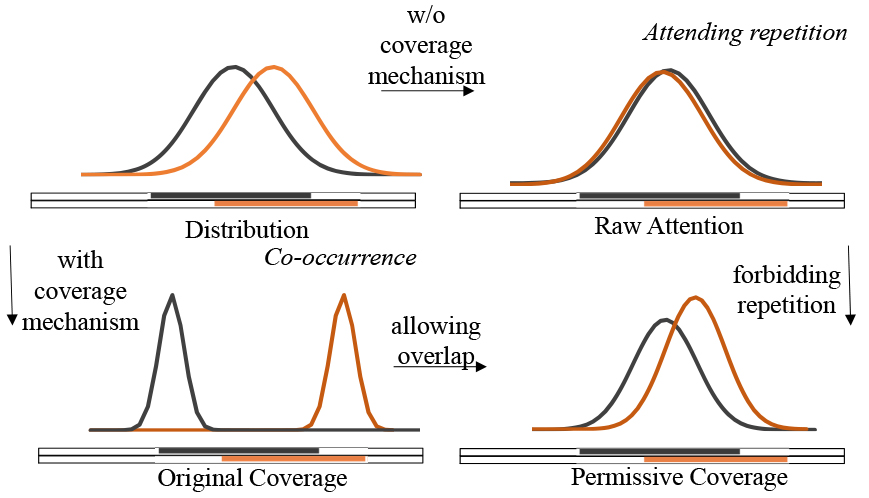}
  \caption{
  Attending weights distribution in different \emph{coverage} mechanisms.
  Horizontal and vertical axes separately stand for time and energy score.
  The \emph{black} and \emph{orange} bars refer to the ground truth of two different actions.
  Curves in \emph{black} and \emph{orange} are corresponding to the energy distributions in two steps.
  }
  \label{fig:cov}
\end{figure}
For \emph{Attending repetition}, coverage mechanism \cite{tu2016modeling} has been introduced to minimize the overlapping of attention weights across time steps, assuming that once given high score-weight in one step, the input vector must not be focused in the future steps.
However, shown as \emph{Original Coverage} in Figure \ref{fig:cov}, conventional coverage mechanism strictly forbids focus on the same place, which is not suitable for video task because of action \emph{co-occurrences} phenomenon.
Instead, we design an action-friendly \emph{permissive coverage} constraints on the weights, in which values of a certain step not only are constraint to the previous weights, but also refer to the last hidden state, which is shown as \emph{Permissive Coverage} in Figure \ref{fig:cov}.
Specifically, we rewrite the coverage score at $t$-th step for segment $s_i$ as
\begin{equation} \label{att5}
\begin{split}
COV_{t,i}&=f(h_{t-1},  \alpha_{t-1,i} ) \\
       &=\sigma[Z_{i}(i-\sum_{k=1}^{N}\alpha_{t-1,k}k) + W_{\alpha}h_{t-1}],
\end{split}
\end{equation}
where $Z_i$ is the corresponding learnable parameter, and \textit{f} is a nonlinear activation function composed of a MLP structure and the \textit{sigmoid} activation.
The ${COV}_{t,i}$ simultaneously restricts the current weight $\alpha_{t}$ attending relevant to $\alpha_{t-1}$ and refers to the last recurrent hidden state ${h}_{t-1}$.

Then we put the coverage term into the attending mechanism by rewriting Equation \ref{att3} as
\begin{equation} \label{att3_ad}
\begin{split}
\hat{e}_{t,i}=&v_{\alpha}[COV_{t,i} \varphi(W_{\alpha}h_{t-1} + U_{\alpha}s_{i})], \\
\end{split}
\end{equation}
where ${COV}_{t,i}$ can be considered as a \textit{soft gating value} within the range of [0,1].
\subsubsection{Repetition Alignment Mechanism}
For \emph{trivial action missing}, we propose a repetition alignment module (RAM) to calculate the frequency of single-instance occurring in each video, which heuristically tunes the action proposal generation. Namely, RAM not only manifests trivial actions occurring, but also restrains the unrelated time segments. 
The RAM at $t$-th step is calculated by
\begin{equation} \label{cnt}
\begin{split}
RAM_{t}&= {W}_{r}\sum_{i=1}^{N}\sigma(\hat e_{t,i}),
\end{split}
\end{equation}
where ${W}_{r}$ is the learned parameters.
$RAM_{t}$ is supervised with the number of corresponding action frequency.
Then we further involve this term into the RNN structures for enhancing the pattern-label relation learning.
Thus Equation \ref{att2} is extended as
\begin{equation} \label{att2_ad}
{h}_{t}=RNN(h_{t-1},  y_{t-1}, x_{t}, RAM_{t}).
\end{equation}
Note that, counting the occurrence of each action category need neither the frame-level information nor precise time locations.
RAM is an effective and flexible assistant sub-module in the whole action assembly generation.

\subsection{Action Proposal Generation}
The Class Activation Mapping (CAM) \cite{Zhou2015Learning} is useful for action localization, and also has been applied in the previous work \cite{Nguyen2017Weakly}.
However, CAM is only designed for linear architectures, and not suitable for nonlinear architectures, such as RNN.
While Gradient-CAM (Grad-CAM) \cite{gradcam} is applicable to any differentiable architecture even with activations.
In this work, we adopt a more general Grad-CAM to calculate class response for our task, termed as Segregated Temporal Gradient-weighted Class Activation Map (ST-GradCAM).

At the $t$-th step, the prediction output $d_t^c$ (output distribution for a class \textit{c} before the \emph{softmax}) is represented by
\begin{equation} \label{cam}
\begin{split}
d_t^c&=\sum_{k=1}^K w_{t,k}^cx_{t}^{k}, \\
\end{split}
\end{equation}
where $w_{t,k}^c$ is the importance of the $k$-th feature value $x_{t}^{k}$ for a target class \textit{c}, which is represented by the following gradient score
\begin{equation} \label{gradcam}
\begin{split}
w_{t,k}^c=&\frac{\partial {d_t^c}}{\partial {x_{t}^k}}
=\frac{\partial {d_{t} ^{c}   }  }{\partial {h_t^c}}\cdot{\frac{\partial {h_t^c}}{\partial {x_{t}^k}}}
={W}_o\cdot{\frac{\partial {{RNN(x_{t})}}}{\partial {x_{t}^k}}},
\end{split}
\end{equation}
where $h_t^c$ is the importance of $h_t$ for the target class \textit{c}.
Since the attention weights possess rich information regarding action intervals \cite{wang2017untrimmednets,Nguyen2017Weakly},
we formulate Equation \ref{cam} as
\begin{equation} \label{cam1}
\begin{split}
d_t^c
=\sum_{k=1}^K w_{t,k}^c(\sum_{i=1}^N\alpha_{t,i}s_{i}^{k}) =\sum_{i=1}^{N}\alpha_{t,i}\sum_{k=1}^K w_{t,k}^cs_{i}^{k},
\end{split}
\end{equation}
where $s_{i}^{k}$ is the ${k^{th}}$ feature value $s_{i}^{k}$ in $s_i$.

Since $d_t^c$ indicates the importance of representations to each class at recurrent step \textit{t}, a class-aware activation map can be derived from above. We define ST-GradCAM as
\begin{equation} \label{cam2}
\begin{split}
\xi_{t,i}^c&= \sum_{k=1}^K w_{t,k}^cs_{i}^{k}, \\
\end{split}
\end{equation}
where $i$ indexes the segment in $S$.
ST-GradCAM captures the important local information of feature map \textit{k} for a target class \textit{c} at recurrent step \textit{t}.

To generate temporal action proposals, we train a two-stream network and derive the attended ST-GradCAM at $t$-th step using $\alpha_{t,i}\cdot \sigma(\xi_{t,i}^c)$.
For each class \textit{c} at the recurrent step \textit{t}, each proposal [$N_{start},N_{end}$] is assigned a score by:
\begin{equation} \label{twostream}
\sum_{i=N_{start}}^{N_{end}}
\frac{ [\lambda\cdot \alpha_{t,i,RGB}^c + (1-\lambda)\cdot \alpha_{t,i,flow}^c] }
{N_{end}-N_{start}+1}\cdot \sigma(\xi_{t,i}^c),
\end{equation}
where we fuse the attention values of RGB and optical flow streams by the modality ratio $\lambda$ ($\lambda=0.5$ by default) first, and then generate proposals based on RGB and flow separately. For final detection, we perform non-maximum suppression (NMS) among temporal proposals of each class by removing highly overlapped ones.

\subsection{Network Training}
The training objective of the STAR network is to solve a multi-task optimization problem.
The overall loss function consists of four terms: the classification loss, coverage loss, repetition alignment loss and sparsity loss,
\begin{equation} \label{loss}
L=L_{class}+\beta L_{sparsity}+\gamma L_{cov} + \delta L_{ram},
\end{equation}
where $\gamma$, $\delta$ and $\beta$ are the hyper-parameters.

Given encoded segment inputs $S$, \textit{classification loss} is defined as the softmax loss over multiple categories by
\begin{equation} \label{clsloss}
\begin{split}
L_{class}&=
         -\frac{1}{M}\sum_{i=1}^{M}\sum_{t=1}^T\log{P_i(\hat{y}_t|S,\theta)},\\
\end{split}
\end{equation}
where $M$, $\hat{y}_t$ and $\theta$ represent the number of training videos, the ground truth of the $t$-th action category, and all the trainable parameters respectively.
$P_{i}$ is the multinomial logistic regression (a probability density over all action categories).

The \textit{coverage loss} is to overcome the common \textit{laziness} of learning problems and thus to put emphasis on different action segments, which is computed by
\begin{equation} \label{covloss}
L_{cov}=max(0,\sum_{i=1}^{N}(\sum_{k=1}^{i}\alpha_{t,k} - \sum_{k=1}^{i}\alpha_{t-1,k} )).
\end{equation}

The \textit{RAM loss} is designed to relieve the \emph{trivial action missing} problem by checking the repetition number, which adopts the L2 loss and is defined as
\begin{equation} \label{ramloss}
L_{ram}= \frac{1}{2T} \sum_{t=1}^{T} || c_t - RAM_{t} ||_2^2,
\end{equation}
where $c_t$ is the ground-truth of the $t$-th action frequency. 

The \textit{sparsity loss} $L_{sparisity}$ is the L1 regularization on the attention weights, i.e., $||\alpha||_1$.

\section{Experiments} \label{expriment}
We evaluate our proposed framework (STAR) with mean average precision (mAP) score on two benchmarks for temporal action detection, i.e., THUMOS'14 \cite{THUMOS14} and ActivityNet1.3 \cite{heilbron2015activitynet:}.
Following the routine evaluation protocol in \cite{Nguyen2017Weakly}, our method outperforms existing weakly-supervised methods. 

\subsection{Implementation Details}
\subsubsection{Datasets.}
\textit{THUMOS'14}, extracted from over 20 hours of sport videos, consists of 20 action classes. It contains 200 videos from validation set for training, and 212 videos for testing.
This dataset is challenging for temporal detection because (1) averagely, each video contains more than 15 occurrences of all actions, (2) the length of an action varies significantly (e.g., from less than one second to over 26 minutes), and (3) averagely, each video is with about 71\% background.
It is a good benchmark for multiple action detection.

\textit{ActivityNet1.3} contains 200 activity categories, in which 10,024 videos are used for training, 4,926 for validation, and 5,044 for testing.
\textit{ActivityNet1.3} only contains 1.5 occurrences per video on average and most videos simply contain single action category with averagely 36\% background.
\subsubsection{Training Details.}
Our model is implemented on Caffe.
For a direct and fair comparison, we follow the video preprocessing procedure of STPN \cite{Nguyen2017Weakly} by pre-training the two-stream I3D network \cite{carreira2017quo} on Kinetics dataset \cite{zisserman2017the}.
Then we uniformly sample 400 segments from each video for feature extraction. The whole network is trained by using Adam optimizer with learning rate $10^{-4}$ and dropout ratio 0.8 on both streams.
Besides, $\beta,\gamma$ and $\delta$ in Equation \ref{loss} are empirically set to be $10^{-4}$, $10^{-4}$ and $10^{-6}$ respectively. 
\subsubsection{Testing Details.}
We retrieve one-dimensional temporal proposals from the predicted label distribution $d$ based on the outputs of the two-streams network.
As the two streams have similar classification performance, we set a modality ratio of 1:1 (RGB:flow) as classification confidence scores and make the prediction jointly.
Then we exploit Equation \ref{twostream} to output the action proposals.

\subsection{Ablation Study}
To analyze the contributions of several different components of STAR, we conduct the ablation study on the THUMOS'14 dataset.
Performance is evaluated with average mAP (\%) by calculating the multiple overlap IoU with thresholds varying from 0.1 to 0.5.
\begin{figure*}[t]
\centering
	\includegraphics[width=0.89\textwidth]{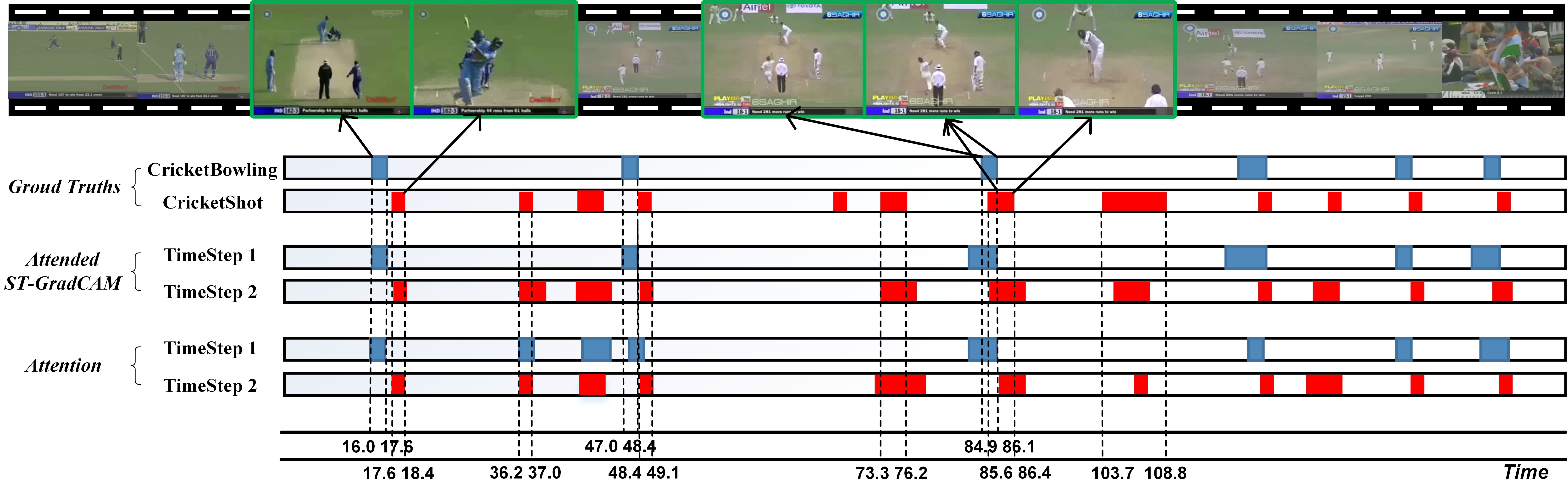}
	\caption{
        An example for action localization on THUMOS'14, which contains two actions (\textit{CricketBowling} denoted and \textit{CricketShot}).
        The video is segregated into two assemblies (TimeStep1 and TimeStep2) step-by-step.
        }
	\label{figure:exp}
\end{figure*}
  \begin{figure*}[!htb]
  	\centering
  	\includegraphics[width=0.89\textwidth]{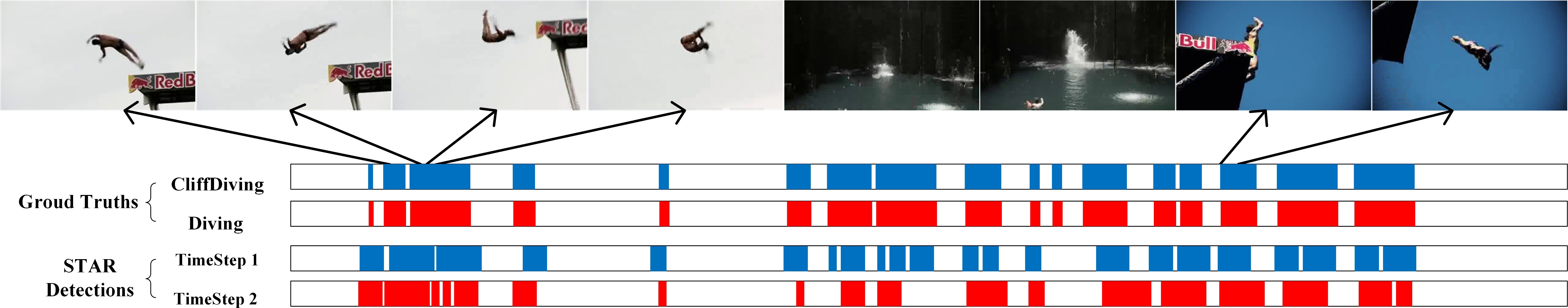}
  	\caption{
  		Results on videos persistently occurring multi-actions.
  	}
  	\label{figure:dense}
  \end{figure*}
  \begin{figure*}[!htb]
  	\centering
  	\includegraphics[width=0.89\textwidth]{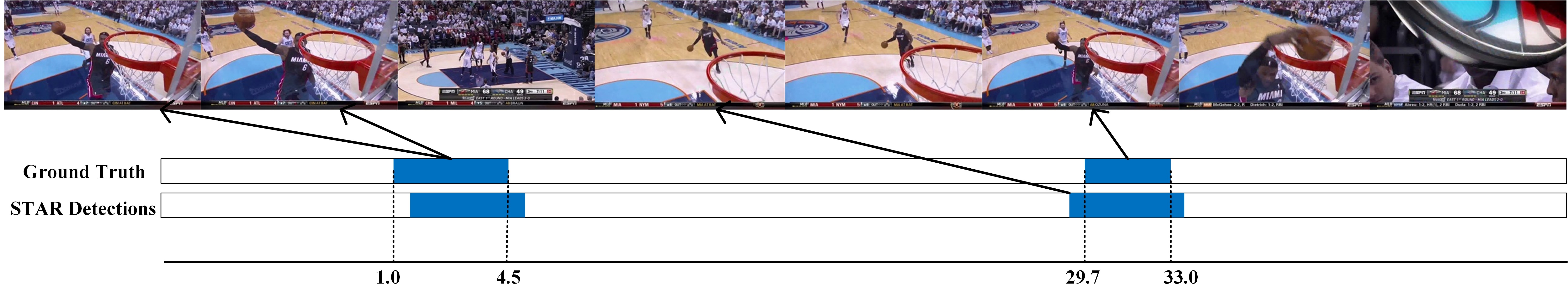}
  	\caption{
  		Results on videos with sparse single-actions.
  	}
  	\label{figure:sparse}
  \end{figure*}
\subsubsection{Effects of architecture modules}
We investigate modules including \textit{coverage} constraints ($COV$), \textit{sparsity} ($SPA$), and \textit{RAM} with \textit{stem} network of STAR. Table \ref{ablation:1} shows the effects of each module and their combinations.
\begin{itemize}
  \item \emph{Stem network}. It serves as our baseline model.
  \item \emph{Stem with one module}. In overall, each single module can improve the stem structure by 3\%-5\%.
      Interestingly, both $(SPA)$ and $(RAM)$ are constraints on the frequency and extent of attended weights, and though by different means, they obtain similar performance alone, better than $(COV)$. We can infer that direct penalty of occurrence is more effective than the handling of action co-occurrence.

  \item \emph{Stem with two modules}.
        RAM with either \emph{sparsity} or \textit{coverage} can achieve better performance than other components, which indicates that additional repetition information is very useful. $(SPA, COV)$ is unsatisfying compared to other module combinations, even worse than $(SPA)$ solely.
        We analyze the reason that both of them restrains the extents of attended weights, but $(RAM)$ aligns the extents by zooming in and out. So combinations with $(RAM)$ achieve steady improvements, while the combination without $(RAM)$ tends to suppress excessively.

  \item \emph{Stem with all modules}. STAR with all modules achieves the best performance which improve the \emph{stem} by 8\%.
\end{itemize}
\begin{table}[h]
	\centering
	\caption{Performance evaluation with different modules of STAR on THUMOS'14.}
\scalebox{0.78}{
	\begin{tabular}{l|c|ccc|ccc|c}
		\hline
		Stem     &  $\surd$  & $\surd$ &  $\surd$ &  $\surd$ &  $\surd$&$\surd$&$\surd$&$\surd$  \\
		Sparsity     &          & $\surd$ &          &         &  $\surd$ &  $\surd$   & 	 &$\surd$  \\
		Coverage     &          &         & $\surd$  &         & $\surd$& &$\surd$ & $\surd$ \\
		RAM       &     &         &          & $\surd$ &      &$\surd$  &   $\surd$    &$\surd$  \\ \hline
		Ave-mAP(\%)      &  39.0   &    43.8     &    42.4      & 43.8 &  43.3    & 44.0  &   44.7   & \textbf{47.0}  \\ \hline		
	\end{tabular}
}
	\label{ablation:1}
\end{table}
\begin{table}[h]
	\centering
	\caption{Effects of different modules on THUMOS'14.}
\scalebox{0.9}{
	\begin{tabular}{l|c}
		Weakly-supervised Model                           &  Ave-mAP(\%)  \\ \hline
        Wang et al. \shortcite{wang2017untrimmednets}   &  29.0         \\
        Singh et al. \shortcite{Singh2017Hide}         &  20.6         \\
        Nguyen et al. \shortcite{Nguyen2017Weakly}      &  35.0         \\
   	   Paul et at.\shortcite{Paul_2018_ECCV}       &  39.7         \\ \hline
        ST-GradCAM                              &  24.4         \\
        Attention                             &  {39.6}         \\
        Attended ST-GradCAM                                      &  \textbf{47.0}  \\ \hline
	\end{tabular}
}
	\label{ablation:2}
\end{table}

\subsubsection{Effects of detection operations}
As introduced in Equation \ref{cam1}, ST-GradCAM and the learned \emph{assembly weights} are respectively responsible for indication of class-specific contribution and input segmented duration.
Since the learned attention weights are sensitive to action classes, we can alternatively use only attention weights without ST-GradCAM to propose locations of each class, termed by \textit{Attention}.
Similarly, the location of each class also can be proposed based on only ST-GradCAM without attention weights,(i.e., \textit{ST-GradCAM}).
We also integrate both learned attention weights and ST-GradCAM as a fused action detector, denoted by \textit{Attended ST-GradCAM} (used by default in our work).

Table \ref{ablation:2} gives the results.
We find that $Attention$ already has achieved comparable performance to previous methods, implying that the learned attention weights themselves contain rich location information and play important roles in the entire detection process.
As expected, the \emph{Attended ST-GradCAM} significantly outperforms each single term (\emph{i.e}, \emph{Attention} and \emph{ST-GradCAM}), which demonstrates the effectiveness of the STAR scoring mechanism (in Equation \ref{cam}).

\subsection{Qualitative Evaluation}
STAR can iteratively segregate different actions from the origin input video segments, then assemble the corresponding actions into a target action-patterns.
For further demonstrating the performance of STAR, we qualitatively analyze the effects of STAR from different aspects as follows:
\begin{itemize}
  \item \emph{Effect of Attention}: The learnt assembly weights are used to assemble actions into the corresponding instance-patterns so that the weights are capable of indicating action locations (\emph{e.g.,} see intervals $[16.0s, 17.6s]$ and $[17.6s, 18.1s]$ in Figure \ref{figure:exp}), in consonance with the performance in subsection \textit{ablation study}. However, it still suffers from ambiguous boundaries of actions (\emph{e.g.} see time $48.4s$ or $85.6s$ in Figure \ref{figure:exp}).
  \item \emph{Effect of Attended ST-GradCAM}: Considering the response mechanism of ST-GradCAM, the \emph{Attended ST-GradCAM} can achieve more precise action location than \emph{Attention} (\emph{e.g.} at time $85.6s$ or $103.7s$ in Figure \ref{figure:exp}).

  \item \emph{False Positive Analysis}: We also analyze the false positives, and find that those falsely detected image frames usually bear high similarities to the annotations, (\emph{e.g.} actions before time $84.9s$ or after time $86.4s$ in Figure \ref{figure:exp}) and are even ambiguous to human beings.

  \item \emph{Evaluation of Extreme Scenarios}: In general, action may occur sparsely or densely in videos, and the degree to which a video is filled with actions can be measured by the action occurring \emph{density}, which is defined as the overlap of all action intervals over the given whole video.
 For example, $denisty=0$ means that there is no actions occurring in the video, while $denisty=1$ means actions occur continuously in the whole video.
  Methods for action detection easily fail in videos with excessively either sparse or dense action occurrence.
  Figure \ref{figure:dense} and \ref{figure:sparse} display the action detection results on \textit{dense} and \textit{sparse} situations respectively,
  which further demonstrate the robustness of our method.
  We also report the performances in terms of Ave-mAP with different densities of occurring actions in videos, which is shown in Figure \ref{figure:quant}. The results show that STAR maintains high performances under different levels of action occurring density.
  \begin{figure*}[!htb]
  	\centering
  	\includegraphics[width=0.85\textwidth, height=0.31\textwidth]{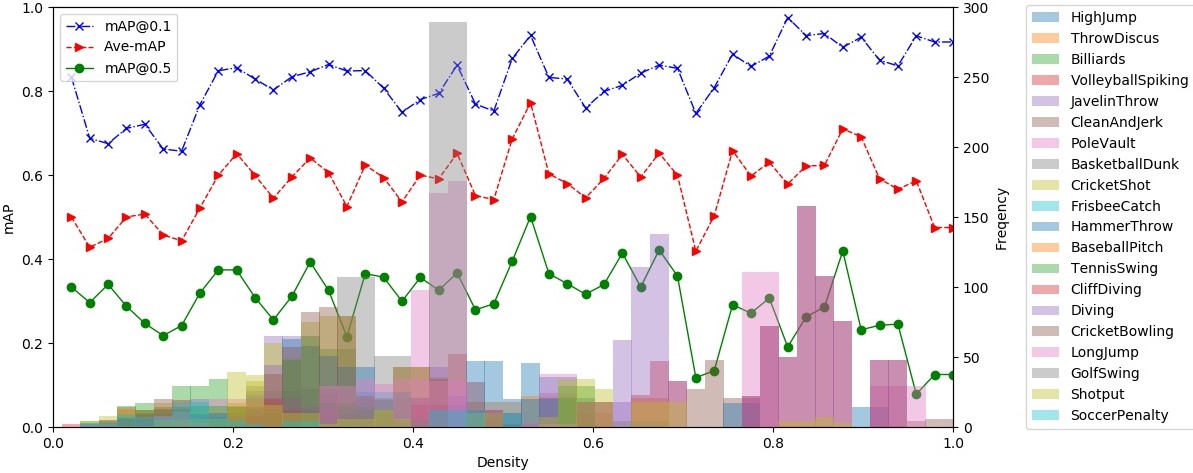}
  	\caption{
  		The performance (mAP)  with varying action \emph{density} values on THUMOS'14.
  		The horizontal, left-vertical and right-vertical axes separately represent the density of action occurring, the mAP and the frequency of action occurring.
  		Three curves describe the mAP performances with IoU 0.1 and IoU 0.5, and the average IoU over 5 thresholds within $[0.1,0.5]$.
  		The colored (denoted as different type of actions) histogram refers to the frequency of corresponding actions occurring at the specific density.
  	}
  	\label{figure:quant}
  \end{figure*}
\end{itemize}

\subsection{State-of-the-Art Comparisons}
We compare STAR with state-of-the-art weakly-supervised and fully-supervised methods on THUMOS'14 and ActivityNet1.3 datasets.
Note that, THUMOS'14 is a better benchmark for evaluating our method, as addressed in \emph{Datasets} section.
Table \ref{comparison:thumos} and \ref{comparison:anet} summarize the results.
\begin{table}[h]
	\centering
	\caption{Comparison with state-of-the-arts on THUMOS'14.
}
\scalebox{0.8}{
	\begin{tabular}{l|l||ccccc}
    \multirow{2}{*}{Supervision} &  \multirow{2}{*}{Method} &
    \multicolumn{5}{c}{AP@IoU} \\
& &  0.1& 0.2 & 0.3 &0.4 &0.5          \\ \hline
\multirow{11}{*}{ \tabincell{c}{Fully \\ Supervised} }
& Richard \shortcite{Richard2016Temporal} &39.7& 35.7 & 30.0 &23.2 &15.2 \\
& Shou \shortcite{Shou2016Temporal}   &47.7& 43.5 & 36.3 &28.7 &19.0  \\
& Yeung \shortcite{yeung2016end-to-end}   &48.9& 44.0 & 36.0 &26.4 &17.1 \\
& Yuan \shortcite{yuan2016temporal}    &51.4& 42.6 & 33.6 &26.1 &18.8 \\
& Shou \shortcite{Shou2017CDC}    & -- &  --  & 40.1 &29.4 &23.3 \\
& Yuan \shortcite{Yuan2017Temporal}    &51.0& 45.2 & 36.5 &27.8 &17.8 \\
& Gao \shortcite{Gao2017Cascaded}    &54.0& 50.9 & 44.1 &34.9 &25.6 \\
& Xu \shortcite{Xu2017R}     &54.5& 51.5 & 44.8 &35.6 &28.9 \\
& Zhao \shortcite{Zhao2017Temporal}    &\textbf{66.0}& \textbf{59.4} & 51.9 &41.0 &29.8 \\
& Lin \shortcite{Lin2017Single}    & 50.1 & 47.8 & 43.0 & 35.0 & 24.6  \\
& Yang \shortcite{yang2018exploring}     & -- & -- & 44.1 & 37.1 & 28.2  \\
& Chao \shortcite{Chao2018Rethinking}    &59.8& 57.1 & 53.2 &\textbf{48.5} &\textbf{42.8} \\
& Alwasssel \shortcite{Alwassel2018ECCV} &49.6& 44.3 & 38.1 &28.4 &19.8 \\
& Lin \shortcite{Lin2018BSN}           &--  & --   & \textbf{53.5} &45.0 &36.9 \\\hline\hline
\multirow{4}{*}{\tabincell{c}{Weakly \\ Supervised}  }
& Wang \shortcite{wang2017untrimmednets}    &44.4& 37.7 & 28.2 &21.1 &13.7 \\
& Singh \shortcite{Singh2017Hide}&36.4& 27.8 & 19.5 &12.7 &6.8 \\
& Nguyen \shortcite{Nguyen2017Weakly}  &{52.0}& {44.7} & {35.5} &{25.8} &{16.9}  \\ 
& Shou \shortcite{Shou_2018_ECCV}  & -- & -- & 35.8 & 29.0 &21.2  \\
& Paul \shortcite{Paul_2018_ECCV}  & {55.2} & {49.6} & 40.1 & 31.1 &{22.8}  \\ \cline{2-7}
& Ours          &\textbf{68.8}& \textbf{60.0} & \textbf{48.7} & \textbf{34.7}&\textbf{23.0}\\
    \hline
	\end{tabular}
}
\label{comparison:thumos}
	
\end{table}
\begin{table}[h]
	\centering
	\caption{A comparison on ActivityNet v1.3 validation set. The sign (*) indicates results from ActivityNet Challenge.
}
\scalebox{0.9}{
	\begin{tabular}{l|l||ccc}
		\multirow{2}{*}{Supervision} &  \multirow{2}{*}{Method} &
		\multicolumn{3}{c}{AP@IoU} \\
		& &  0.5& 0.75 & 0.95  \\ \hline
		\multirow{6}{*}{\tabincell{c}{Fully \\ Supervised} }
		& Singh \shortcite{Singh2016Untrimmed}* & 34.5 & --   & -- \\
		& Shou \shortcite{Shou2017CDC}* & 45.3 & 26.0       & 0.2     \\
		& Dai \shortcite{Dai2017Temporal}*   & 36.4 & 21.2       & 3.9     \\
		& Xiong \shortcite{Xiong2017A}* & 39.1 & 23.5      & 5.5     \\
		& Lin \shortcite{Lin2017Single}*  & 49.0 & 32.9 &  7.9    \\
		& Xu \shortcite{Xu2017R}   & 26.8 &  --        &  --     \\
		& Chao \shortcite{Chao2018Rethinking} & 38.2 & 18.3       & 1.3     \\
		& Lin \shortcite{Lin2018BSN} &\textbf{52.5}& \textbf{33.5}& \textbf{8.9}  \\\hline\hline
		\multirow{2}{*}{\tabincell{c}{Weakly \\ Supervised}  }
		& Nguyen \shortcite{Nguyen2017Weakly} & {29.3} & {16.9}    & {2.6}     \\
		& Ours       & \textbf{31.1} & \textbf{18.8}    & \textbf{4.7}     \\
		\hline
	\end{tabular}
}
	\label{comparison:anet}
\end{table}

\emph{Comparison with weakly supervised methods}.
It is shown that STAR outperforms all other weakly supervised methods by a remarkably large margin, improving the reported highest average mAP about {7\% by 47.0\% on THUMOS'14 and 2\% by 18.1\%} on ActivityNet1.3. {Note that, on THUMOS'14, our model achieves more than 10\% higher than existing methods on both IoU 0.1 and 0.2. This verifies the superiority of our framework.
}

\emph{Comparison with fully supervised methods}.
In Table \ref{comparison:thumos}, our model also has comparable results with those fully supervised methods.
There is a great gap between the existing fully and weakly supervised methods because of the usages of detailed boundary annotations in fully supervised methods. However, our model still outperforms all fully supervision results at both IoU 0.1 and 0.2 on THUMOS'14 dataset.

As in Table \ref{comparison:anet}, results with asterisk (*) are collected from ActivityNet Challenge submissions (only for general references here), which can not be impartially compared directly with our STAR.
{
We see that our model even overtakes the recent strong-supervised method \cite{Xu2017R} and partially surpasses method \cite{Chao2018Rethinking}, but falls behind work BSN \cite{Lin2018BSN}. Note that, BSN takes full use of boundary annotations via a sophisticated multi-stage training strategy.}
In conclusion, STAR surpasses all the reported weakly-supervised methods on both two benchmarks. Although fully-supervised approaches still have good results at large IoU thresholds, STAR significantly narrows down the gap between the fully and weakly supervised methods.

\section{Conclusion}
We propose an end-to-end weakly supervised framework STAR for action detection in MIML perspective.
The model first assembles actions into corresponding instance-patterns with a well-designed attention mechanism, and then learns the temporal relationship between multiple instance-patterns by using RNN.
Finally, with the predicted action labels and the learned attention weights, we use a well designed ST-GradCAM for localizing each action.
Experiments show that our approach outperforms all the reported results by weakly supervised approaches by a large margin, and also achieves comparable performance with those fully supervised methods on both THUMOS'14 and ActivityNet1.3 datasets.

\fontsize{9.4pt}{10.4pt}
\selectfont
{\small
\bibliographystyle{aaai}
\bibliography{aaairef}
}

\end{document}